%% file: main.tex
\RequirePackage{xcolor}
\relax
\documentclass[letterpaper]{article} 

\usepackage[final]{corl_2019}  
\usepackage{times}  
\usepackage{helvet} 
\usepackage{courier}  
\usepackage{graphicx} 
\usepackage{graphicx}  
\usepackage{algorithm}
\usepackage{algorithmic}
\usepackage{amsfonts}       
\usepackage{amsmath} 
\usepackage{amssymb} 
\usepackage{amsthm}
\usepackage{booktabs} 
\usepackage{enumitem}
\usepackage{epstopdf} 
\usepackage{float}
\usepackage[T1]{fontenc}    
\usepackage{footmisc} 
\usepackage{hypernat}
\usepackage[utf8]{inputenc} 
\usepackage{mathtools}
\usepackage{microtype} 
\usepackage{multirow}
\usepackage{nicefrac}  
\usepackage{smartref}
\usepackage{stackengine}
\usepackage{subcaption}
\usepackage{todonotes}
\usepackage{xcolor}
\usepackage[justification=centering]{caption}

\newcommand{\cA}{\mathcal{A}}

\newcommand{\cD}{\mathcal{D}}

\newcommand{\cL}{\mathcal{L}}

\newcommand{\cN}{\mathcal{N}}

\newcommand{\cP}{\mathcal{P}}

\newcommand{\cS}{\mathcal{S}}

\newcommand{\Real}{\mathbb{R}}

\newcommand{\Esp}{\mathop{\mathbb{E}}}

\newcommand{\Normal}{\cN}

\newcommand{\new}{\text{new}}
\newcommand{\old}{\text{old}}
\newcommand{\kl}{\text{D}_{\text{KL}}}

\definecolor{myred}{RGB}{219, 48, 122}

\renewcommand{\vec}[1]{\ensuremath{\mathbf{#1}}}

\newcommand{\mat}[1]{\ensuremath{\mathbf{#1}}}

\title{Leveraging exploration in off-policy algorithms via normalizing flows}

\author{
  \Large Bogdan Mazoure\thanks{These authors contributed equally.}\textsuperscript{\quad 1,2}, \Large  Thang Doan\textsuperscript{\text{*} }\textsuperscript{1,2}, \Large  Audrey Durand\textsuperscript{2,3},\\
  {\bf \Large Joelle Pineau\textsuperscript{1,2,4}, R Devon Hjelm\textsuperscript{2,5,6}} 
  \\\textsuperscript{1}McGill University$\qquad$\textsuperscript{2}Mila -- Quebec AI Institute$\qquad$\textsuperscript{3}Universit\' e Laval
  \\\textsuperscript{4}Facebook AI Research$\qquad$\textsuperscript{5}Microsoft Research Montreal
  \\\textsuperscript{6}Universit\' e de Montr\' eal}

\begin{document}
\maketitle

\begin{abstract}
The ability to discover approximately optimal policies in domains with sparse rewards is crucial to applying reinforcement learning (RL) in many real-world scenarios.  Approaches such as neural density models and continuous exploration (e.g., Go-Explore) have been proposed to maintain the high exploration rate necessary to find high performing and generalizable policies.
Soft actor-critic (SAC) is another method for improving exploration that aims to combine efficient learning via off-policy updates, while maximizing the policy entropy.
In this work, we extend SAC to a richer class of probability distributions (e.g., multimodal) through normalizing flows (NF) and show that this significantly improves performance by accelerating discovery of good policies while using much smaller policy representations.
Our approach, which we call SAC-NF, is a simple, efficient, easy-to-implement modification and improvement to SAC on continuous control baselines such as MuJoCo and PyBullet Roboschool domains. Finally, SAC-NF does this while being significantly parameter efficient, using as few as $5.5\%$ the parameters for an equivalent SAC model.
\end{abstract}

\section{Introduction}

\input{introduction}

\section{Related Work}
\label{sec:rel_work}
\input{related}

\section{Background}
\label{sec:background}
\input{background}

\section{Augmenting SAC with Normalizing Flows}
\label{sec:sac_nf}
\input{sac_nf.tex}

\section{Experiments}
\label{sec:experiments}
\input{experiments.tex}

\section{Conclusion}
\label{sec:conclusion}

We proposed an algorithm which combines soft actor-critic updates together with a sequence of normalizing flows of arbitrary length. The high expressivity of the later allows to (1) quickly discover richer policies (2) compress the cumbersome Gaussian policy into a lighter network and (3) better avoid local optima. Our proposed algorithm leverages connections between maximum entropy reinforcement learning and the evidence lower bound used to optimize variational approximations. Finally, we validated the model on six MuJoCo tasks, three Bullet Roboschool tasks and one sparse domains, on which SAC-NF showed significant improvement against the SAC baseline in terms of convergence rate as well as performance. Interesting challenges for future work include studying the generalization and theoretical properties of normalizing flow SAC policies to better transfer from rich simulators to real robots.


\section*{Acknowledgements}
We want to thank Compute Canada/Calcul Qu\' ebec and Mila -- Quebec AI Institute for providing computational resources. We also thank Chin-Wei Huang for insightful discussions.

\newpage
\bibliographystyle{aaai}

\input{main.bbl}
\newpage
\onecolumn

\section*{Supplementary Material}
\input{supplementary.tex}

\end{document}

%% file: introduction.tex
Reinforcement learning (RL) provides a principled framework for solving continuous control problems, yet current RL algorithms often do not explore well enough to solve high-dimensional robotics tasks~\cite{ecoffet2019go}. Environments with a large number of continuous control factors, such as those that involve combinations of leg movement, arm movement, posture, etc, have many local minima \cite{lehmanPHD}. For example, it is possible to achieve forward momentum in humanoid environments \cite{lehmannovelty} with a variety of suboptimal policies over those factors (e.g. arms lean forward or backward, not synchronized with legs), in a way that will fail readily as the environmental variables change (such as environments designed to purposely destabilize the agent, e.g., \citeauthor{coumans2016pybullet} \citeyear{coumans2016pybullet}).
Success in these environments requires a complex coordination of the control factors, and to learn this, it is necessary to have an exploration strategy that avoids converging too early on suboptimal local minima~\cite{Conti2018Improving}.

Soft Actor-Critic~(SAC)~\cite{sac} is a state-of-the-art exploration-based algorithm that adds a maximum entropy bonus term~\cite{williams1991function} to a differentiable policy objective specified by a soft critic function.
As an off-policy algorithm, SAC enjoys sample efficiency -- a desirable property in robotics, where real world experiments might be costly to perform.
However, SAC is limited to modeling policies that have closed-form entropy (e.g., unimodal Gaussian policies), which we posit hurts exploration~\cite{haarnoja2017reinforcement}.  The main contribution of this work is to extend SAC to a richer class of multimodal exploration policies, by transforming the actions during exploration via a sequence of invertible mapping known as \textit{normalizing flows} (NF)~\cite{rezende2015variational}. 
Our approach, which we call SAC-NF, is a simple and easy-to-implement extension to the original SAC algorithm that gives the agent access to a more expressive multimodal policy and that achieves much better performance on continuous control tasks.

We show empirically that this simple extension significantly improves upon the already high exploration rate of SAC and achieves better convergence properties as well as better performance on both sparse and deceptive environments. Next, the class of policies that we propose requires significantly less parameters than its baseline counterpart, while also improving on the baseline results. Finally, we assess the performance of both SAC and SAC-NF across a variety of benchmark continuous control tasks from OpenAI Gym using the MuJoCo simulator~\cite{mujoco} and the realistic Bullet Roboschool tasks~\cite{coumans2016pybullet}.

%% file: related.tex
\paragraph{Off-policy RL}
Off-policy strategies in RL collect samples under some behaviour policy and use those samples to train a target policy. 
Off-policy algorithms are known to train faster than their on-policy counterparts, but at the cost of higher variance and instability~\cite{ddpg}. 
Among this family, actor critic (AC) strategies have shown great success for solving continuous control tasks. In between value-based and policy-based approaches, an AC algorithm trains an \textit{actor} (policy-based) using guidance from a \textit{critic} (value-based). Two major AC algorithms, SAC~\cite{sac} and TD3~\cite{td3}, have shown a large performance improvement over previous off-policy algorithms such as DDPG~\cite{ddpg} or A3C~\cite{A3C}. TD3 achieved this by maintaining a second critic network to alleviate the overestimation bias, while SAC enforced more exploration by adding an entropy regularization term to the loss function.

\paragraph{Density estimation for better exploration}
Using powerful density estimators to model state-action values with the aim to improve exploration generalization has been a long-standing practice in RL. For instance, \cite{alpha_div_policy_gradient}
use dropout approximation~\cite{gal2016dropout} within a Bayesian network and show improvement on stability and performance of policy gradient methods. \cite{boostrapped_dqn} rather rely on an ensemble of neural networks to estimate the uncertainty in the prediction of the value function, allowing to reduce learning times while
improving performance. Finally, \cite{gan_dqn} consider generative adversarial networks~\cite{gan} to model the distribution of random state-value functions. The current work considers a different approach based on normalizing flows for density estimation.

\paragraph{Normalizing flows}
Flow-based generative models have proven to be powerful density approximators~\cite{rezende2015variational}. The idea is to relate an initial noise density distribution to a posterior distribution using a sequence of invertible transformations, parametrized by a neural network and having desirable properties. For example, invertible autoregressive flows (IAF) are characterized by a simple-to-compute Jacobian~\cite{iaf}. In their original formulation, IAF layers allow learning location-scale invariant (i.e. affine) transformations of a simple initial noise density.

Normalizing flows have been used previously in the on-policy RL setting where IAF extends a base policy found by TRPO~\cite{boosting_trpo}. In this work, we tackle the off-policy learning setting, and we focus on planar and radial flows, which are known to provide a good trade-off between function expressivity and time complexity~\cite{rezende2015variational}.
Our work explores a similar space as hierarchical-SAC (HSAC)~\cite{haarnoja2018latent}, which also modifies the policy of SAC to improve expressiveness. However, HSAC has a significantly more complex model, as it uses real NVP~\cite{dinh2016density} along with a hierarchical policy, optimizing a different reward function at each hidden layer.
This represents a stronger departure from the original SAC model and algorithm. We show that simply training all NF layers jointly on a single reward function without any additional conditioning produces significant improvement over SAC and HSAC with a model and training procedure that is reasonably close to SAC.

%% file: background.tex
In this section, we review the formal setting of RL in a Markov decision process (MDP), Soft Actor-Critic~(SAC)~\cite{sac}, and the general framework of normalizing flows~(NFs)~\cite{rezende2015variational}, the latter of which will be used to improve exploration in Section~\ref{sec:sac_nf}.

\subsection{Markov Decision Process}

MDPs~\cite{bellman1957markovian,puterman2014markov} are useful for modelling sequential decision-making problems.
A discrete-time finite-horizon MDP
is described by a state space $\cS$, an action space $\cA$\footnote{The state $\cS$ and the action $\cA$ spaces can be either discrete or continuous}, a transition function $\cP:\cS\times \cA \times \cS\mapsto \Real^+$, and a reward function $r:\cS\times \cA\mapsto \Real$.
On each round $t$, an agent interacting with this MDP observes the current state $s_t\in\cS$, selects an action $a_t\in\cA$, and observes a reward $r_t = r(s_t, a_t) \in\Real$ upon transitioning to a new state $s_{t+1}\sim \cP(s_t, a_t)$.
Let $\gamma\in[0,1)$ be a discount factor;
the goal of an agent evolving in a discounted MDP is to learn a policy $\pi:\cS\times\cA\mapsto[0,1]$, such that taking action $a_t\sim\pi(\cdot|s_t)$ would maximize the expected sum of discounted returns,
\begin{align*}
V^\pi(s) = \Esp_\pi \bigg[\sum_{t=0}^\infty \gamma^t r_t | s_0 = s\bigg].    
\end{align*}
The corresponding state-action value function can be written as the expected discounted rewards from taking action $a$ in state $s$, that is,
\begin{align*}
Q^\pi(s,a)=\Esp_{\pi}\bigg[\sum_{i=t}^\infty\gamma^{i-t}r(s_i,a_i)|s_t=s,a_t=a\bigg].
\end{align*}
If $\cS$ or $\cA$ are vector spaces, action and space vectors are respectively denoted by $\vec{a}$ and $\vec{s}$.

\subsection{Soft Actor-Critic}

SAC~\cite{sac} is an off-policy algorithm which updates the policy using gradient descent, minimizing the KL divergence between the policy and the Boltzmann distribution using the critic (i.e., Q-function) as a negative energy function,
\begin{align}
    \pi_\new =\arg\min_{\pi'\in \Pi} \kl\bigg(\pi'(.|\vec{s}_t) \bigg\lVert \frac{\exp\{\frac{1}{\alpha}Q^{\pi_\old}(\vec{s}_t,.)\}}{Z^{\pi_\old}(\vec{s}_t)} \bigg),
    \label{eq:sac_pi_loss}
\end{align}
where $\alpha\in (0,1)$ controls the temperature, $Q^{\pi_\old}$ is the  Q-function under the old policy, and the partition function $Z^{\pi_\old}(\vec{s}_t)$ can be ignored~\cite{sac}. 
The KL divergence above is tractable and differentiable as the policies are assumed to be composed of diagonal Gaussians in the classical SAC formulation.
It can be seen easily that SAC follows a maximum entropy objective~\cite{williams1991function}, as optimizing w.r.t. Equation~\ref{eq:sac_pi_loss} is equivalent to maximizing the state-action value function regularized with a maximum entropy term,
\begin{align}
    \label{eqn:max_entropy_objective}
    \begin{split}
        \cL_\pi &= \Esp_{s_t\sim \rho_\pi} \left[\Esp_{a_t\sim \pi}[Q^\pi(s_t,a_t)]+\alpha H(\pi(.|s_t))\right]\\
    &=\Esp_{\vec{s}_t \sim \rho_\pi}[V(\vec{s}_t) ]; \nonumber\\
    &\text{where} \quad V(\vec{s}_t) := \Esp_{a_t\sim \pi}[Q^\pi(s_t,a_t) - \alpha \log \pi(a_t|s_t)]
    \end{split}
\end{align}
is the state-action value function, $H(\pi(.|s_t))$ is the entropy of the policy, $\alpha$ is now the importance given to the entropy regularizer. 
If $\pi'(\cdot|\vec{s}_t)\sim\mathcal{N}(\vec{\mu},\text{diag}(\sigma^2_1,..,\sigma^2_d))$, then $\max H(\pi'(\cdot|\vec{s}_t))=\max  \log\det(\text{diag}(\sigma^2_1,..,\sigma^2_d))=\max \sum_{i=1}^d\sigma^2_i$, which is unbounded without additional constraints. This prevents collapse to degenerate policies centered at the point with highest rewards and keeps exploration active.\\
SAC models the value function and the critic using neural networks, $V_\nu$ and $Q_\omega$, and models a Gaussian policy $\pi_\theta$ with mean and variance determined by the output of neural networks with parameters $\theta$.
The losses for $V_\nu$, $Q_\omega$, and $\pi_\theta$ are computed using a replay buffer $\mathcal{D}$,
\begin{align}
    \cL_Q &=\Esp_{(\vec{s}_t,\vec{a}_t)\sim \cD}\bigg[\frac{1}{2}\big\{Q_\omega(\vec{s}_t,\vec{a}_t)
    -Q^{\dagger}_\nu\big\}^2\bigg],\\
    \cL_V &= \Esp_{\vec{s}_t \sim \cD}\bigg[\frac{1}{2}\big\{V_{\nu}(\vec{s}_t)  -\Esp_{\vec{a}_t\sim \pi}[Q_\omega(\vec{s}_t,\vec{a}_t)-\alpha\log\pi(\vec{a}_t|\vec{s}_t)]\big\}^2\bigg],\\
    \cL_\pi &= \Esp_{\vec{s}_t\sim \cD}\left[\Esp_{\vec{a}_t\sim\pi}[\alpha \log \pi_{\theta,\phi}(\vec{a}_t|\vec{s}_t)-Q_\omega(\vec{s}_t,\vec{a}_t)]\right],
\end{align}
where $Q^{\dagger}_\nu=(r(\vec{s}_t,\vec{a}_t) +\gamma \Esp_{\vec{s}_{t+1}\sim \cD}[V_{\nu}(\vec{s}_{t+1})])$.\\
In practice, the gradients of the above losses are approximated using Monte-Carlo.
As an off-policy algorithm, SAC enjoys the advantages of having lower sample complexity than on-policy algorithms~\cite{trpo}, yet it outperforms other off-policy alternatives~\cite{ddpg} due to its max entropy term.



\subsection{Normalizing Flows}

NFs~\cite{rezende2015variational} are a class of iterative methods for transforming probability distributions introduced as a way to improve the approximate posterior in amortized inference algorithms~\cite{salimans2015markov,hjelm2016iterative}.
More generally, they provide a framework for extending the change of variable theorem for density functions to a sequence of $d$-dimensional real random variables $\{\vec{z}_i\}_{i=0}^N$. The initial random variable $\vec{z}_0$ has density function $q_0$ and is linked to the final output of the flow $\vec{z}_N$ through a sequence of invertible, smooth mappings $\{f_i\}_{i=1}^{N}$ called \textit{normalizing flows} of length $N$. A number of different invertible function families can be specified through the choice of neural network parameterization and regularization~\cite{rezende2015variational,iaf,naf}.
One good choice~\cite{iaf} is the family of radial contractions around a point $\vec{z}_0\in\Real^d$ defined as~\cite{rezende2015variational},
\begin{align}
    f(\vec{z})=\vec{z}+\frac{\beta}{\alpha+||\vec{z}-\vec{z}_0||_2}(\vec{z}-\vec{z}_0),
\end{align}
which are highly expressive (i.e. represent a wide set of distributions) and yet very light (parameter-wise), in addition to enjoying a closed-form determinant.
This family allows approximating the target posterior through a sequence of concentric expansions of arbitrary width and centered around a learnable point $\vec{z}_0$. In order to guarantee that the flow is invertible, it is sufficient to pick $\beta\geq -\alpha$.

%% file: sac_nf.tex
We now propose a flow-based formulation of the off-policy maximum entropy RL objective (Eq.~\ref{eqn:max_entropy_objective}) and argue that SAC is a special case of the resulting approach, called SAC-NF.

\subsection{Exploration through normalizing flows}
\begin{figure}
    \centering
    \includegraphics[width=0.45\linewidth]{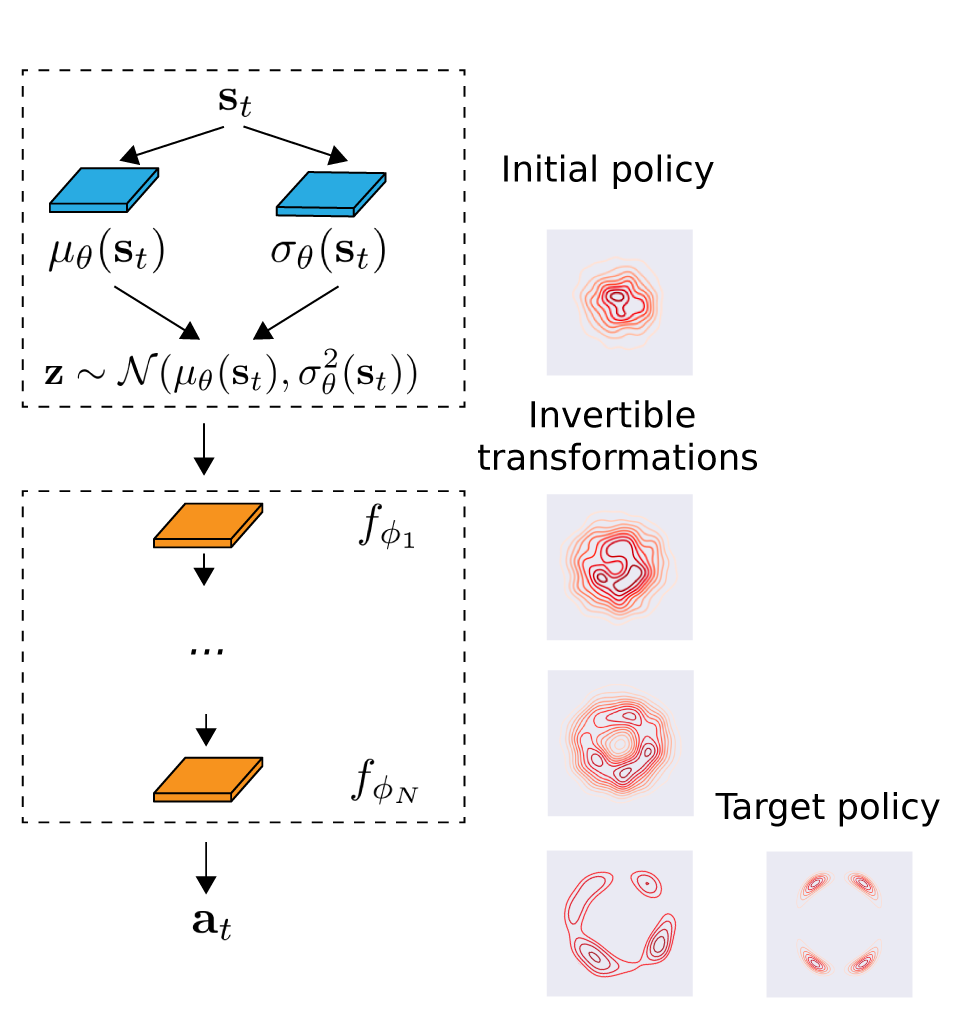}
    \caption{Schematic representation of a SAC-based normalizing flow policy, which takes as input noise and state information. The initial noise density is then fed through a sequence of invertible transformations to match the target Boltzmann Q-function.}
    \label{fig:cartoon}
\end{figure}

Figure~\ref{fig:cartoon} outlines the architecture of a normalizing flow policy based on SAC.
Let $\vec{\varepsilon}$ be an initial noise  sample, $h_\theta(\vec{\varepsilon},\vec{s}_t)$ a state-noise embedding, and $\{f_{\phi}\}_{i=1}^N$ a normalizing flow of length $N$ parameterized by $\phi=\{\vec{\phi}_i\}_{i=1}^N$. Sampling from the policy $\pi_{\phi,\theta}(\vec{a}_t|\vec{s}_t)$ can be described by the following set of equations:
\begin{align}
    &\vec{a}_t = f_{\phi_{N}} \circ f_{\phi_{N-1}} \circ...  \circ f_{\phi_{1}} (\vec{z}), \nonumber\\
    &\vec{z} = h_\theta^{j}(\vec{\varepsilon},\vec{s}_t),  & j=1,2 \nonumber\\
    &\vec{\varepsilon} \sim \mathcal{N}(0,\mat{I}),
\end{align}
where the state-noise embedding $h_\theta^{j}(\vec{\varepsilon},\vec{s}_t)$ models samples from a base Gaussian distribution, with state-dependent means, $\mu_\theta(\vec{s_t})$. The index $j$ denotes a hyperparameter, choosing either state-dependent ($\mat{L}_{\theta}(\vec{s_t})$ for $j = 1$) or state-independent ($\mat{L}_{\theta}$ for $j = 2$) diagonal scale matrices,
\begin{align}
    &h_\theta^j(\vec{\varepsilon}_0,\vec{s}_t)=\begin{cases}
    \vec{\varepsilon} \mat{L}_\theta(\vec{s}_t) + \mu_\theta(\vec{s}_t), & j=1 \text{ (conditional)}\\
    \vec{\varepsilon} \mat{L}_\theta + \mu_\theta(\vec{s}_t), & j=2 \text{ (average)};\\
    \end{cases}\nonumber\\
    \text{where} \quad
    &\mat{L}(\vec{s_t})=\text{diag}\{(\sigma_1(\vec{s}_t),..,\sigma_{|\mathcal{A}|}(\vec{s}_t))\} \nonumber\\
    &\mat{L}=\text{diag}\{(\sigma_1,..,\sigma_{|\mathcal{A}|})\}
\end{align}
We chose $j$ according to experiments in the supplementary.
Both functions allow to sample either from a heteroscedastic or homoscedastic Gaussian distribution, following the reparametrization trick in variational Bayes~\cite{kingma2013auto}, and we explore these choices in more detail in the supplementary material.
Precisely, $\mu(\vec{s}_t):\mathcal{S}\to \mathbb{R}^d$ is a state embedding function and $\mat{L}(\vec{s}_t),\mat{L}$ is a scale parameter.
For flows of the form $f_\phi(\vec{z})=\vec{z}+g_\phi(\vec{z})$, we can asymptotically recover the original base policy through heavy regularization,
\begin{align}
    \lim_{||\vec{\phi}_1||,..,||\vec{\phi}_N||\to 0}\pi(\cdot|\vec{s}_t)\overset{d}{=}\Normal\big(\mu(\vec{s}_t),\mat{L}(\vec{s}_t)\mat{L}(\vec{s}_t)^\top\big),
\end{align}
for all states $\vec{s}_t\in \mathcal{S}$.
By analogy with the SAC updates, SAC-NF minimizes the KL divergence between the Boltzmann $Q$ and the feasible set of normalizing flow-based policies. The KL term is once again tractable and the policy density now depends on the sum of log Jacobians of the flows:
\begin{align}
    \log\pi(\vec{a}_t,\vec{s}_t)
    =\log q_0(\vec{a}_0)
    &- \log|\det\mat{L}| \nonumber\\
    &-\sum_{i=1}^N\log \Bigl| \det \frac{\partial f_i(\vec{a}_{i-1})}{\partial \vec{a}_{i-1}} \Bigr|.
\end{align}
Algorithm~\ref{alg:sac_nf} outlines the proposed method: the major distinction from the original SAC is the additional gradient step on the normalizing flows layers while fixing the SAC weights $\theta$.
\begin{algorithm}[ht]
    \caption{SAC-NF}
    \label{alg:sac_nf}
    \begin{algorithmic}
       \STATE {\bfseries Input:} Mini-batch size $m$; replay buffer $\cD$; number of epoch 
       $T$;
       learning rates $\alpha_{\theta},\alpha_{\phi},\alpha_{\nu},\alpha_{\omega}$\\
       \STATE Initialize value function network $V_\nu(\vec{s})$
       \STATE Initialize critic network $Q_\omega(\vec{s},\vec{a})$
       \STATE Initialize policy network with weights $\pi_{\phi,\theta}(\vec{s})$
       \FOR{$\text{epoch}=1,...,T$} 
       \STATE $  \vec{s}  \gets \vec{s}_0$
        \FOR{t=0...}
       \STATE $\vec{a}_t \sim \pi(.|\vec{s}_t)$
       \STATE Observe $\vec{s}_{t+1} \sim P(\cdot|\vec{s}_t,\vec{a}_t)$ and get reward $r_t$
       \STATE Store transition ($\vec{s}_t$,$\vec{a}_t$,$r_t$,$\vec{s}_{t+1}$) in $\cD$
       
       \FOR{each learning step}
       \STATE \algorithmiccomment{Update networks with $m$ MC samples each}
        \STATE $  \nu  \gets \nu - \alpha_{\nu}  \nabla_{\nu} \hat{\mathcal{L}}_{V}$ \algorithmiccomment{Update value function}
        \STATE $  \omega  \gets \omega - \alpha_{\omega}   \nabla_{\omega} \hat{\mathcal{L}}_{Q}$ \algorithmiccomment{Update critic}
        \STATE $  \theta  \gets \theta - \alpha_{\theta}  \nabla_{\theta} \hat{\mathcal{L}}_{\pi}$ \algorithmiccomment{Update base policy}
        \STATE $  \phi  \gets \phi - \alpha_{\phi}  \nabla_{\phi} \hat{\mathcal{L}}_{\pi}$ \algorithmiccomment{Update NF layers}
       \ENDFOR
       \ENDFOR
       \ENDFOR
    \end{algorithmic}
\end{algorithm}


%% file: experiments.tex
This section addresses three major points: (1) it highlights the beneficial impact of NF deceptive rewards domains through a navigation task, (2) compares the proposed SAC-NF approach against SAC on a set of continuous benchmark control tasks from MuJoCo, Rllab and the more realistic Roboschool PyBullet suite~\cite{mujoco,duan2016benchmarking,coumans2016pybullet} and finally (3) investigates the shape and number of modes that can be learned by radial NF policies. 
For all experiments\footnote{We trained all policies on Intel Gold 6148 Skylake @ 2.4 GHz processors.}, the entropy rate $\alpha$ is constant and can be found in Supp.~Table~\ref{tab:experiments_parameters}. For the SAC baseline, we used hyperparameters reported in \cite{sac}.
A thorough study of multi-modality and non-Gaussianity of SAC-NF policies on MuJoCo is shown in the Appendix.


\subsection{Robustness to confounding rewards}

\begin{figure}
    \centering
    \includegraphics[width=0.6\linewidth]{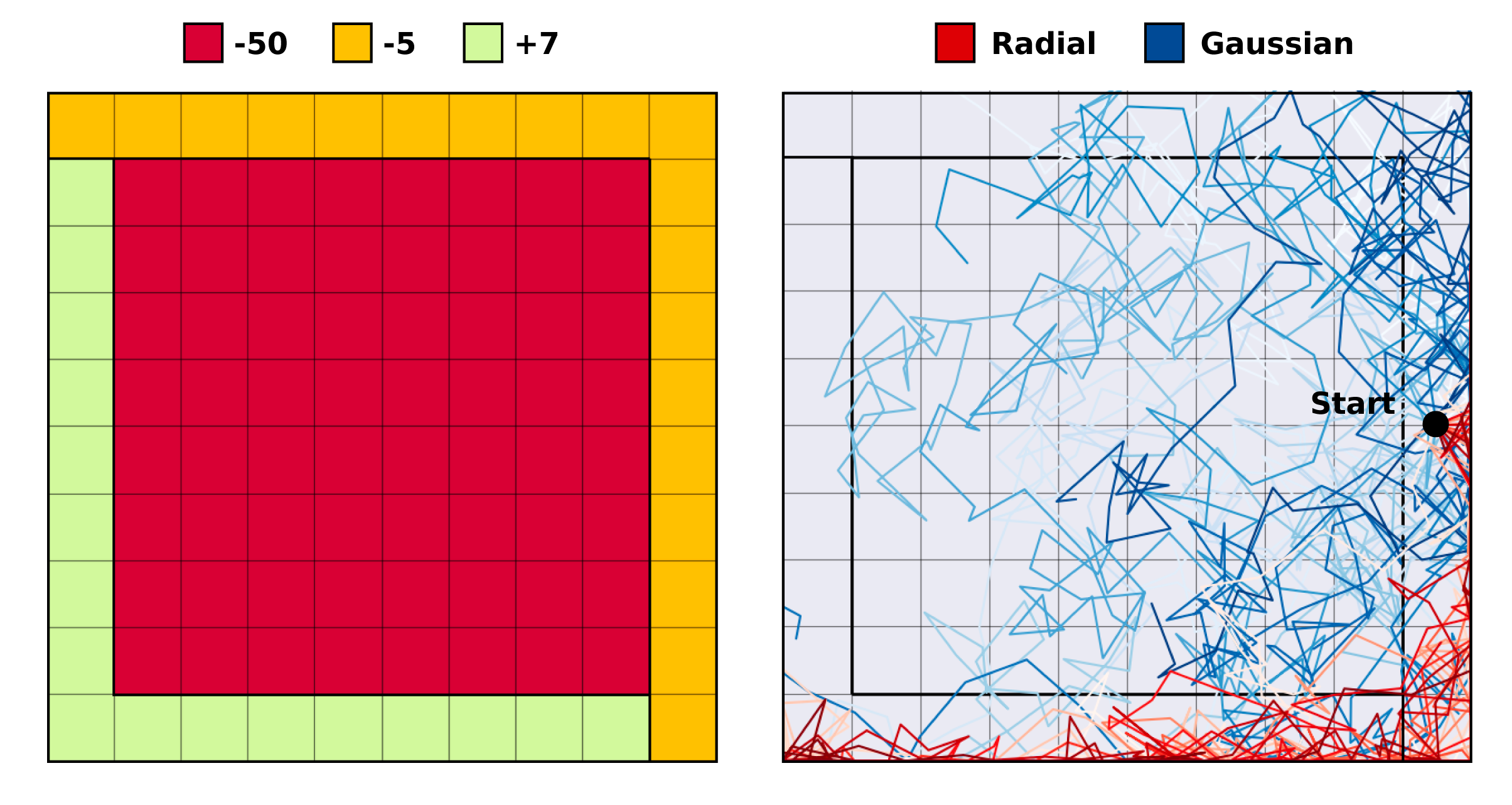}
    \caption{A deceptive reward environment where the high reward region is hidden behind a negative reward region (left subfigure). The right subfigure shows trajectories of Gaussian and radial agents. While the SAC policy is exploring vaguely the yellow region and falls into the pit, the SAC-NF policy manages to find the optimal region by going from the yellow to green zone.}
    \label{fig:pitfall_env}
\end{figure}

We first demonstrate that normalizing flow policies are able to find better solutions than a Gaussian policy for SAC in an environment with deceptive rewards.
We consider an environment composed of three reward areas: a locally optimal strip around the initial state, a global optimum on the opposing end of the room, separated by a pit of highly negative reward. The agent starts at the position $\Vec{s}_0 = (4.5,0)$ and must navigate into the high rewards area without falling into the pit. On each time $t$, the agent receives the reward $r_t$ associated to its current location $\Vec{s}_t$. The experimental setup can be found in Supplementary Material.

Figure~\ref{fig:pitfall_env} displays the trajectories visited by both agents. This highlights the biggest weakness of vanilla SAC policies: the agent is unable to simultaneously reach the region of high rewards while avoiding the center of the room. In this case, lowering the entropy threshold will lead to the conservative behaviour of staying in the yellow zone; increasing the entropy leads the agent to die without reaching the goal. Breaking the symmetry of the policy by adding (in this case three) radial flows allows the agent to successfully reach the target area by walking along the safe path surrounding the room.

In the case of steep reward functions, where low rewards border on high rewards, symmetric distributions force the agent to explore into all possible directions. This leads the agent to sometimes attain the high reward region, but, more dangerously, falling into low reward areas with non-zero probability at training time.


\subsection{Continuous control tasks}

\subsubsection{MuJoCo locomotion benchmarks}

\begin{figure*}
    \centering
    \includegraphics[width=0.8\linewidth]{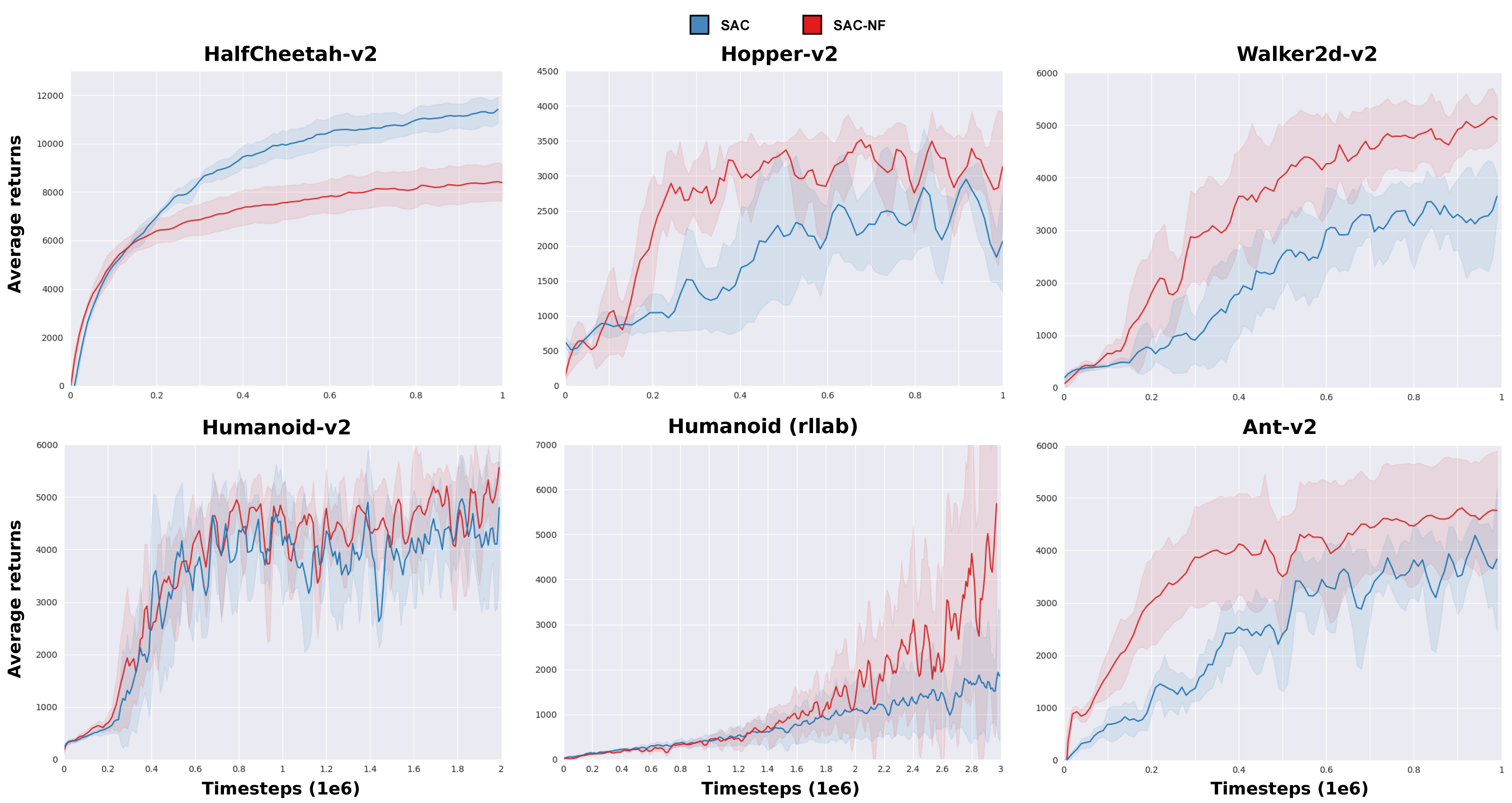}
    \caption{Performance of SAC-NF against SAC across 6 MuJoCo tasks (higher is better). Curves are averaged over 5 random seeds and then smoothed using Savitzky-Golay filtering with window size 7.}
    \label{fig:mujoco_curves}
\end{figure*}

Next, we compare our SAC-NF method against the SAC baseline on six continuous control tasks from the MuJoCo suite (see Figure~\ref{fig:mujoco_curves}) and one sparse reward MuJoCo task\footnote{The sparse Humanoid task can be found here: \url{https://github.com/bmazoure/sparseMuJoCo}}. All results curves show evaluation time performance which, in the case of SAC and SAC-NF, is equivalent to setting the noise to 0. Evaluation happens every 10,000 steps, and values reported in the tables are not smoothed. The values reported in the plots are smoothed with a window size of 7, equivalent to smoothing every 70,000 steps to improve readability.


The SAC-NF agent consists of one feed-forward hidden layer of $256$ units acting as state embedding, which is then followed by a normalizing flow of length $N\in \{3,4,5\}$. Details of the model can be found in Supp.~Table~\ref{tab:experiments_parameters}. For the SAC baseline, two hidden layers of $256$ units are used. The critic and value function architectures are the same as in \cite{sac}. All networks are trained with Adam optimizer~\cite{adam} with a learning rate of $3\text{E}^{-4}$. Every $10,000$ environment steps, we evaluate our policy $10$ times and report the average. The best observed reward for each method can be found in Table~\ref{tab:reported_rewards}. 

Figure~\ref{fig:mujoco_curves} displays the performance of both SAC and SAC-NF. We observe that SAC-NF shows faster convergence, which translates into better sample efficiency, compared to the baseline. SAC-NF takes advantage of the expressivity of normalizing flows to allow for better exploration and thus discover new policies. In particular, we notice that SAC-NF performs well on three challenging tasks: \texttt{Humanoid-v2}, \texttt{Humanoid (rllab)} and \texttt{Ant-v2}. High rewards collected by SAC-NF agents suggest that Gaussian policies that are widely used for continuous control~\cite{ppo,trpo} might not be best suited for certain domains (see Supplementary Material for a shape analysis of SAC-NF policies on \texttt{Ant-v2}).

Table~\ref{tab:reported_rewards} not only shows better performance from SAC-NF in most of the environments, but shows the ratio in the number of parameters in the policy architecture between SAC-NF and vanilla SAC. For instance, on \texttt{Hopper-v2}, we could reduce by up to $95\%$ the number of parameters ($70,406$ parameters for SAC baseline versus $3,861$ for SAC-NF) and by $41\%$ the number of parameters in \texttt{Humanoid-v2}, while performing at least as well as the baseline. For space constraints, we also reported results from TD~\cite{td3} in the Supplementary material.


\begin{table}[h!]

    \centering
    {\small \begin{tabular}{c|c|c|c}
    \hline
        &SAC & SAC-NF &  $\frac{\#\{\text{SAC-NF}\}}{\#\{\text{SAC}\}}$   \\
    \hline
     Ant-v2 & $4,372 \pm 900$ & \textbf{4912 $\pm$ 954}  & $\approx 0.31$ \\
     HalfCheetah-v2& \textbf{11410 $\pm$ 537} & 8429 $\pm$ 818 & $\approx 0.09$  \\
    Hopper-v2 & $3095 \pm 730$ & \textbf{3538 $\pm$ 108} & $\approx 0.06$ \\
     Humanoid-v2& \textbf{5505 $\pm$ 116} & \textbf{5506 $\pm$ 147} & $\approx 0.6$\\
     Humanoid (rllab) & $2079 \pm 1432$ & \textbf{5531 $\pm$ 4435} & $\approx 0.4$\\
     Walker2d-v2  & $3813 \pm 374$ & \textbf{5196 $\pm$ 527}  & $\approx 0.09$ \\
     SparseHumanoid-v2  & 88 $\pm$ 159 &  \textbf{547 $\pm$ 268} & $\approx 0.6$  \\
         \hline
    \end{tabular}}
    \vspace{.5em}
    \caption{Maximal average return $\pm$ one standard deviation across 5 random seeds for SAC and SAC-NF. Last column shows the ratio in number of policy parameters between the two methods. Learning curve for SparseHumanoid-v2 can be found in the Appendix.}
    \label{tab:reported_rewards}
\end{table}

SAC could be run with fewer parameters for better comparison with the small architecture of SAC-NF. We also run SAC with a reduced number of hidden units (64 and 128 for the policy network only). In general, running SAC with fewer parameters achieves worse results: best results with either 64 or 128 units are as follows (5 seeds, after 1M steps, 3M for Rllab): 7300 versus 11,410 for the 256 units architecture (HalfCheetah), 4400 versus 5505 (Humanoid), 2900 versus 2079 (Humanoid-rllab), 3800 versus 3813 (Ant).

\subsubsection{Realistic continuous control with Bullet Roboschool}

To assess the behaviour of SAC-NF in realistic environments, we tested our algorithm on the PyBullet Gym implementation\footnote{The environment can be found at: \url{https://github.com/benelot/pybullet-gym}} of Roboschool tasks \cite{coumans2016pybullet}. The Bullet library is among the most realistic collision detection and multi-physics simulation engines available up to now, and is widely used for sim-to-real transfer tasks.

\begin{figure}[t]
    \centering
    \includegraphics[width=1\linewidth]{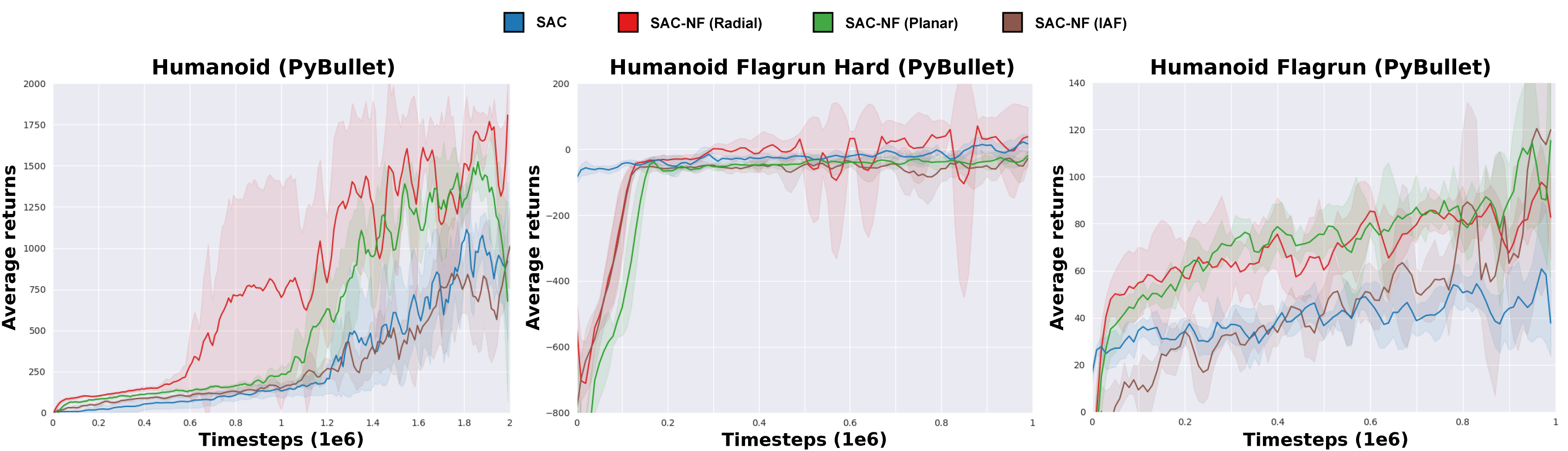}
    \caption{Performance of SAC-NF with IAF, planar, and radial flows compared against SAC (Gaussian policy) accross 3 Roboschool PyBullet domains (higher is better). Curves are averaged over 5 random seeds and then smoothed using Savitzky-Golay filtering with window size 7. Radial flows perform consistently well across the 3 robotics environments.}
    \label{fig:bullet_curves}
\end{figure}

To assess the impact of flow family on performance, we compared three types of normalizing flows: radial, planar, and IAF\footnote{We adapted the more general implementation from: \url{https://github.com/CW-Huang/naf}}. Figure~\ref{fig:bullet_curves} displays the performance of both SAC and SAC-NF for all three flows families obtained using the same setup as for MuJoCo.

The best observed reward for each method can be found in Supp.~Table~\ref{tab:reported_rewards_bullet}. SAC-NF with radial flows consistently ranks better (performance and parameter-wise, see Supp.~Table~\ref{tab:nb_params_pybullet}) than the Gaussian policy and, in some domains, better than planar and IAF flows.


\begin{table*}[t]
    \centering
    {\small \begin{tabular}{c|c|c|c|c|c}
    \hline
        & SAC & Radial  & Planar & IAF  \\
    \hline
     Humanoid (PyBullet) & $1263 \pm 290$ & \textbf{1755 $\pm$ 131} & $1561 \pm 238$ & $1130 \pm 206$  \\
      Humanoid Flagrun Hard (PyBullet) & $31 \pm 33$ &\textbf{46 $\pm$ 61} & $-10 \pm 16$ &  $-2 \pm 33$   \\
    Humanoid Flagrun (PyBullet) & $67 \pm 41$ & $106 \pm 23$ &  $128 \pm 42$ & \textbf{152 $\pm$ 173} \\
   
         \hline
    \end{tabular}}
    \vspace{.5em}
    \caption{Maximal average return obtained on three Roboschool PyBullet environments by Gaussian, IAF, planar, and radial policies 
    $\pm$ one standard deviation across 5 random seeds.}
    \label{tab:reported_rewards_bullet}
\end{table*}

%% file: supplementary.tex
\subsection*{Reproducibility Checklist}

We follow the reproducibility checklist~\cite{joelle_reproduciblity} and point to relevant sections explaining them here.
\\
For all algorithms presented, check if you include:\\
\begin{itemize}
    \item \textbf{A clear description of the algorithm, see main paper and included codebase.}
    The proposed approach is completely described by Alg.~\ref{alg:sac_nf}.
\item \textbf{An analysis of the complexity (time, space, sample size) of the algorithm.}
Experimentally, we demonstrate improvement in sample complexity as discussed in our main paper. In term of computation time, the proposed method retains the same running time as SAC, since the overhead for training the NF layers is minimal in comparison. The biggest advantage of SAC-NF with radial flows over SAC is its significantly reduced number of parameters. For instance, in the MuJoCo Hopper environment, SAC-NF uses only $5.5\%$ of neural network parameters used by SAC while achieving much better performance.

\item \textbf{A link to a downloadable source code, including all dependencies.} The code is included with Supplemental Material as a zip file; all dependencies can be installed using Python's package manager. Upon publication, the code would be available on Github.
\end{itemize}
For all figures and tables that present empirical results, check if you include:
\begin{itemize}
    \item \textbf{A complete description of the data collection process, including sample size.} We use standard benchmarks provided in OpenAI Gym (Brockman et al., 2016) and PyBullet.
    \item \textbf{A link to downloadable version of the dataset or simulation environment.} See: \url{http://www.mujoco.org/} and \url{https://pybullet.org/wordpress/}.
\item \textbf{An explanation of how samples were allocated for training / validation / testing.} We do not use a training-validation-test split, but instead report the mean performance (and one standard deviation) of the policy at evaluation time across 5 random seeds.
\item \textbf{An explanation of any data that were excluded.} We did not compare on easy environments (e.g. \texttt{Reacher-v2}) because all existing methods perform well on them. In that case, the improvement of our method upon baselines is incremental and not worth mentioning.
\item \textbf{The exact number of evaluation runs.} 5 seeds for all experiments, 1M, 2M or 3M environment steps depending on the domain.
\item \textbf{A description of how experiments were run.} See Section~\ref{sec:experiments} in the main paper and didactic example details in Appendix.
\item \textbf{A clear definition of the specific measure or statistics used to report results.} Undiscounted returns across the whole episode are reported, and in turn averaged across 5 seeds.
\item \textbf{Clearly defined error bars.} Confidence intervals and table values are always mean$\pm$ $1$ standard deviation over 5 seeds.
\item \textbf{ A description of results with central tendency (e.g. mean) and variation (e.g. stddev)}. All results use the mean and standard deviation.
\item \textbf{ A description of the computing infrastructure used.} All runs used 1 CPU for all experiments (toy, MuJoCo and PyBullet) with $8$Gb of memory. 
\end{itemize}

\newpage
\subsection*{Experimental setup for ablation study}
We compare the SAC-NF agent (Algorithm~\ref{alg:sac_nf} with mini-batch size $m=256$, 4 flows and one hidden layer of 8 neurons), which can represent radial policies, with a classical SAC agent(two hidden layers of 16 units) that models Gaussian policies. Both agents are trained over $T=500$ epochs, each epoch consisting of $20$ time steps.

\subsection*{Model parameters}
We provide a table of hyperparameters used to obtain results in the MuJoCo and PyBullet domains. Note that $h^1$ corresponds to the average and $h^2$ to the conditional models.

\begin{table}[H]
\centering
\resizebox{0.8\textwidth}{!}{
\begin{tabular}{l|c|c|c|c}
 \cline{1-5}
    \multicolumn{5}{c}{NF parameters}         \\  \hline
                      & $\#$ flows & Type  &  Alpha &  Model \\  \cline{1-5}
\multicolumn{1}{c|}{ Ant-v2}    & $4$ & radial &  $0.05$  &  average  \\ 
\multicolumn{1}{c|}{ HalfCheetah-v2}           & $3$ & radial &  $0.05$  & conditional   \\ 
\multicolumn{1}{c|}{ Hopper-v2 }     &  $5$ & radial &  $0.05$ & average \\  
\multicolumn{1}{c|}{ Humanoid-v2}     & $4$ & radial &  $0.05$ & average  \\   
\multicolumn{1}{c|}{ Walker-v2}     & $5$ & radial &  $0.05$  &  conditional \\
\multicolumn{1}{c|}{ Humanoid (Rllab)}     & $2$ & radial &  $0.05$ & conditional  \\ 
\multicolumn{1}{c|}{ HumanoidPyBulletEnv-v0}     & $3$ & radial &  $0.05$  &  average \\ 
\multicolumn{1}{c|}{ HumanoidFlagrunPyBulletEnv-v0}     & $5$ & radial &  $0.05$  &  conditional \\
\multicolumn{1}{c|}{ HumanoidFlagrunHarderPyBulletEnv-v0}     & $3$ & radial &  $0.05$  &  conditional \\
\multicolumn{1}{c|}{ HumanoidPyBulletEnv-v0}     & $3$ & IAF &  $0.01$  &  conditional \\ 
\multicolumn{1}{c|}{ HumanoidFlagrunPyBulletEnv-v0}     & $4$ & IAF &  $0.05$  &  conditional \\
\multicolumn{1}{c|}{ HumanoidFlagrunHarderPyBulletEnv-v0}     & $3$ & IAF &  $0.01$  &  average \\
\multicolumn{1}{c|}{ HumanoidPyBulletEnv-v0}     & $4$ & planar &  $0.01$  &  conditional \\ 
\multicolumn{1}{c|}{ HumanoidFlagrunPyBulletEnv-v0}     & $3$ & planar &  $0.05$  &  average \\
\multicolumn{1}{c|}{ HumanoidFlagrunHarderPyBulletEnv-v0}     & $3$ & planar &  $0.05$  &  average \\\hline
  \multicolumn{5}{c}{Adam Optimizer parameters} \\  \hline
\multicolumn{1}{c|}{ $\alpha_{\gamma}$}     & \multicolumn{1}{|c}{$3.10^{-4}$}  \\
\multicolumn{1}{c|}{ $\alpha_{\omega}$}     & \multicolumn{1}{|c}{$3.10^{-4}$}  \\
\multicolumn{1}{c|}{ $\alpha_{\theta}$}     & \multicolumn{1}{|c}{$3.10^{-4}$}  \\
\multicolumn{1}{c|}{ $\alpha_{\phi}$}     & \multicolumn{1}{|c}{$3.10^{-4}$}  \\ \hline
 \multicolumn{5}{c}{Algorithm parameters}         \\  \hline
 \multicolumn{1}{c|}{ $m$}     & \multicolumn{1}{|c}{$256$}  \\
 \multicolumn{1}{c|}{ $\mathcal{B}$ size}     & \multicolumn{1}{|c}{$10^6$}  \\ \hline
\end{tabular}}
\caption{SAC-NF parameters.}
\label{tab:experiments_parameters}
\end{table}

\begin{table}[H]
\centering
\begin{tabular}{l|c|c|c|c}
 Environment & Gaussian & Radial & IAF & Planar\\
 \hline
HumanoidPyBulletEnv-v0  &  82,463 ($1$) & 15,963 ($0.19$) & 17,436 ($0.21$) & \textbf{13,594} ($0.16$)\\ 
HumanoidFlagrunPyBulletEnv-v0 & 82,463 ($1$) & \textbf{12,397} ($0.15$) & 18,864 ($0.23$) & 16,875 ($0.20$)\\
HumanoidFlagrunHarderPyBulletEnv-v0 & 82,463 ($1$) & \textbf{12,359} ($0.15$) & 21,040 ($0.26$) & 16,875($0.20$)\\\hline
\end{tabular}
\caption{Number of model parameters for SAC (Gaussian), SAC-NF (Radial, Planar and IAF) used to achieve results on the PyBullet environments. In parentheses, the ratio of parameters with respect to SAC (Gaussian) is shown. A value lower than $1.0$ means a lower number of parameters than SAC baseline. While having the lowest number of parameters, radial flows achieve consistently best performances.}
\label{tab:nb_params_pybullet}
\end{table}

\newpage

\subsection*{Performances against other baselines}

\begin{table}[h!]
    \centering
    {\small \begin{tabular}{c|c|c|c}
    \hline
        &SAC & SAC-NF & TD3    \\
    \hline
     Ant-v2 &  $4,372 \pm 900$ & \textbf{4912 $\pm$ 954} & $4,372 \pm 900$   \\
     HalfCheetah-v2& \textbf{11410 $\pm$ 537} & 8429 $\pm$ 818 & $9,543 \pm 978$    \\
    Hopper-v2 & $3095 \pm 730$ & \textbf{3538 $\pm$ 108}  & \textbf{3,564 $\pm$ 114}  \\
     Humanoid-v2& \textbf{5505 $\pm$ 116} & \textbf{5506 $\pm$ 147} &  $71 \pm 10$ \\
     Humanoid (rllab) & $2079 \pm 1432$ & \textbf{5531 $\pm$ 4435} & $286\pm 151$ \\
     Walker2d-v2  & $3813 \pm 374$ & \textbf{5196 $\pm$ 527} & $4,682 \pm 539$  \\
     SparseHalfCheetah-v2  & $767 \pm 247$ & \textbf{939 $\pm$ 4} & $809 \pm 92$ \\
     SparseHumanoid-v2  & 88 $\pm$ 159 &  \textbf{547 $\pm$ 268} & $0 \pm 0$  \\

         \hline
    \end{tabular}}
    \vspace{.5em}
    \caption{Maximal average return $\pm$ one standard deviation across 5 random seeds for SAC, TD3 and SAC-NF.}
\end{table}

\subsection*{Toy navigation task}

We conduct a synthetic experiment to illustrate how the augmentation of a base policy with normalizing flows allows to represent multi-modal policies. We consider a navigation task environment with continuous state and action spaces consisting of four goal states symmetrically placed around the origin. The agent starts at the origin and, on each time $t$, receives reward $r_t$ corresponding to the Euclidean distance to the closest goal.
We consider a SAC-NF agent (Algorithm~\ref{alg:sac_nf} with mini-batch size $m=256$, 4 flows and one hidden layer of 8 neurons) which can represent radial policies. The agent is trained over $T=500$ epochs, each epoch consisting of $20$ time steps.

\begin{figure}[h!]
    \centering
    \includegraphics[width=0.6\linewidth]{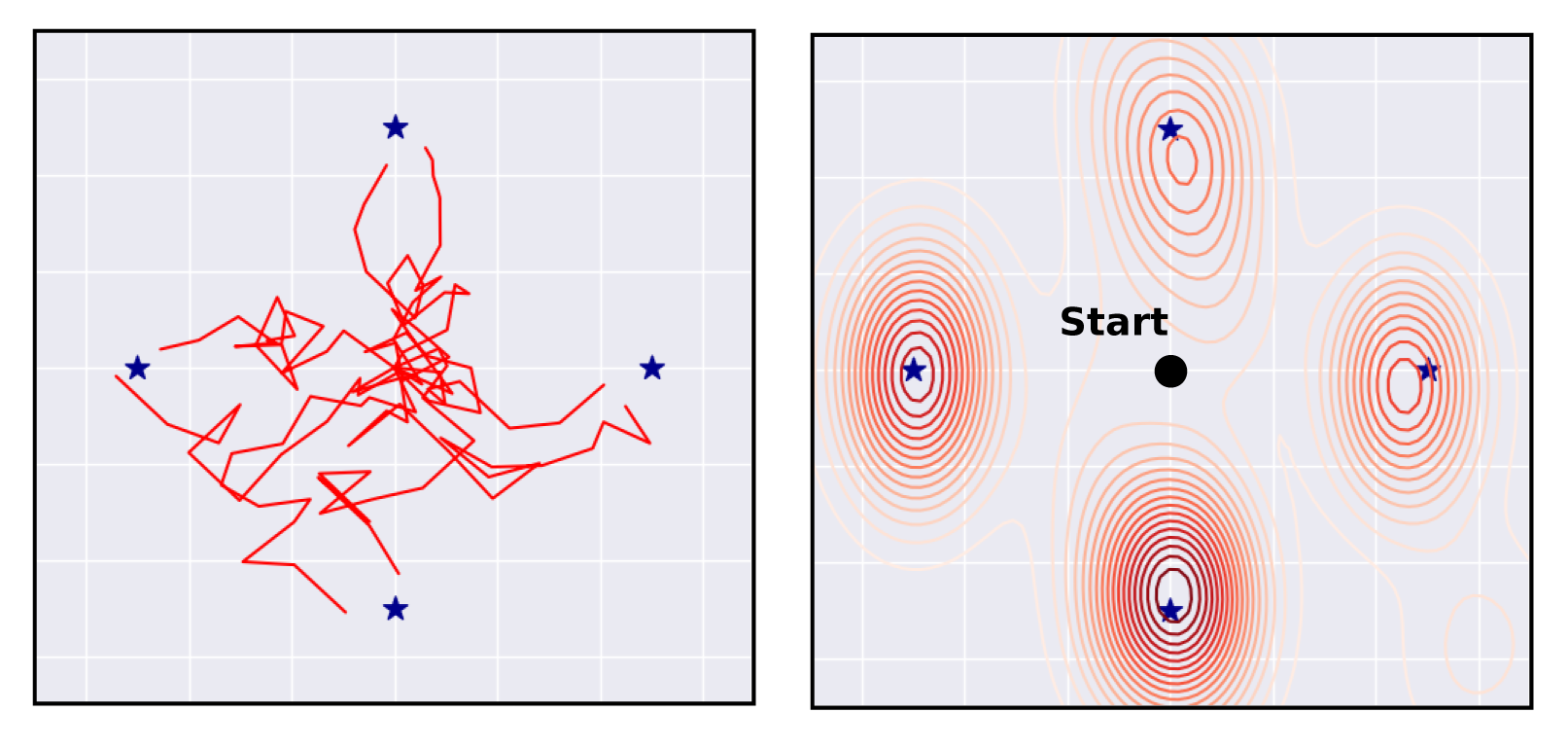}
    \caption{4-goals navigation task outlining the ability of normalizing flows to learn multi-modal policies. The left subfigure shows some trajectories sampled from the SAC-NF agent. The right subfigure shows a KDE plot of terminal state visitations by the agent.}
    \label{fig:4_goal_env}
\end{figure}

Figure~\ref{fig:4_goal_env} displays some trajectories sampled by the SAC-NF agent along with the kernel density estimation (KDE) of terminal state visitations by the agent. Trajectories are obtained by sampling from respective policy distributions instead of taking the average action. We observe that the SAC-NF agent, following a flow-based policy, is able to successfully visit all four modes.

\newpage

\subsection{Assessing the shape of SAC-NF and its multimodality} 
\label{sec:non_gauss}
While in simple domains the shape of a policy might not matter much, using a Gaussian policy in more complicated environments can yield suboptimal behaviour due to rigidity of its shape. To test whether SAC-NF with radial flows implicitly learns a Gaussian policy (i.e. most learning happens at the noise and not at the flow layers), we examine the KL divergence between a Gaussian distribution and SAC-NF policies trained on MuJoCo's \texttt{Ant-v2} environment.\\
As argued previously, heavy regularization of radial and planar flows approximately recovers the identity map $f(\vec{z})=\vec{z}$, in which case the normalizing flow policy has a Gaussian shape centered at $\mu(\vec{s})$. However, when the flows are unconstrained, the policy is allowed to evolve as to maximize the evidence lower bound.

\begin{figure}[h!]
    \centering
    \includegraphics[width=0.6\linewidth]{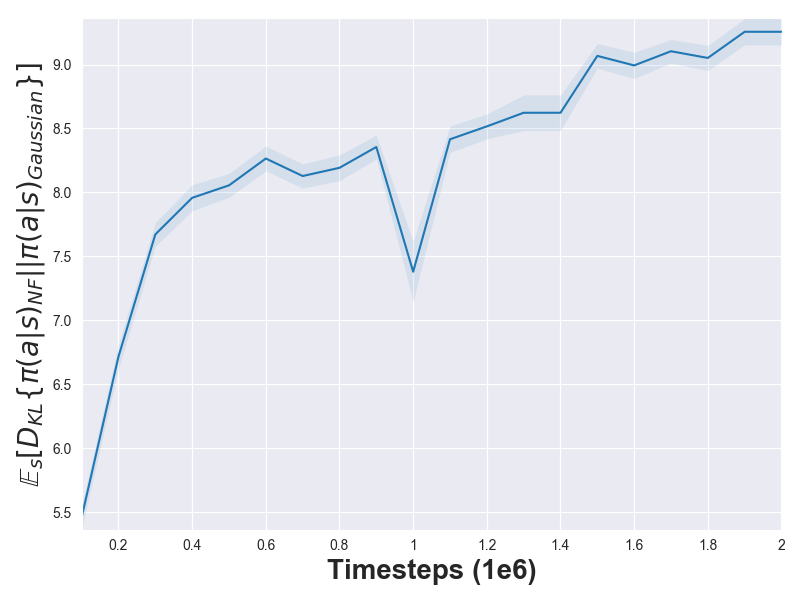}
    \caption{Average KL divergence between 500 MC action samples taken from a radial flow policy scored against the standard Gaussian distribution and averaged over 1,000 states every 100k iterations. The KL divergence increases over the time, suggesting that the radial flow policy's shape gets further from that of a Gaussian distribution.}
    \label{fig:analysis_kl_nf_gaussian}
\end{figure}

Figure~\ref{fig:analysis_kl_nf_gaussian} shows the evolution over time of a radial policy on the MuJoCo \texttt{Ant-v2} environment. The average KL divergence conditional on an observed state is computed between zero mean and unit variance radial flow and Gaussian policies, respectively. This standardization is done to eliminate the dependence of KL on the location and scale of the policy. For two multivariate Gaussian policies $\pi_1\sim \mathcal{N}(\mu_1,\mat{I}),\pi_2\sim \mathcal{N}(\mu_2,\mat{I})$ in a single state environment, the KL divergence follows this proportionality: $D_{KL}(\pi_1||\pi_2)\propto (\mu_1-\mu_2)^\top(\mu_1-\mu_2)$.\\
To ensure that the KL reports the difference in shape and not in location-scale, it is necessary to re-center and re-scale both policies (equivalent to superposing both policies on top of each other):
\begin{equation}
\begin{split}
     &\mathbb{E}_\vec{s}[\text{D}_{KL}\{\pi_{NF}(\vec{a}|\vec{s})||\pi_{Gaussian}(\vec{a}|\vec{s})\}]\\
     &=\mathbb{E}_\vec{s}[D_{KL}\{\pi_{NF}(\vec{a}|\vec{s})||\mathcal{N}(0,\mat{I})\}],
\end{split}
\end{equation}
and $\mathbb{E}_{\pi}[\pi_{NF}(\vec{a}|\vec{s})]=0,\mathbb{V}_{\pi}[\pi_{NF}(\vec{a}|\vec{s})]=1$ for all states $\vec{s}$ observed during rollouts. The differences in log probabilities for every given action are summed over all actions in the sample and averaged across states. We see that as training progresses, the state-averaged KL between the normalizing flow policy and the reference unit Gaussian increases.

Now, we check the key property of SAC-NF, as a proposed improvement to unimodal Gaussian policies in SAC, is its ability to produce rich, multimodal policies. We can measure the degree of multimodality using the gap statistic method \cite{tibshirani2001estimating} for $k$-means hyperparameter selection. Once computed, this coefficient measures goodness-of-fit of $k$ clusters (i.e. modes) to a given distribution. For that purpose, we collect 500 states from the given policy rollout, under which we sample 250 actions from the SAC-NF policy and evaluate the number of modes with the aforementioned test for each state separately. Differences between GS values for $k$=1 and $k$=2, and between $k$=1 and $k$=10 are given respectively and are averaged across all states: 6.8, 11.8 (Ant), 16.8, 31 (Walker), 3.2, 7.3 (Humanoid-rllab). Here, $k$ is the number of clusters, higher difference means more likely to have more than one mode. In comparison, all gap statistics for SAC have differences less than 1. These results suggest that the SAC-NF policies are multimodal (have at least 2 modes, note this is for each state, not marginalized over states). 
Since modes are not always clearly identifiable, we computed skewness (symmetry measure) and excess kurtosis (non-Gaussianity measure), respectively, for the same policy samples: -0.017, -0.26 (Ant), -0.34, -0.9 (Humanoid-rllab), -0.53, -0.57 (Walker). All three policies have large negative excess kurtosis (suggesting that they do not have a Gaussian shape), and have negative skew (policies learned on Humanoid-rllab and Walker are hence not symmetric). This evidence indicates that the shape of the policies learned by SAC-NF is, on average, not likely to be Gaussian.

\subsection{Sparse rewards environments}
\begin{figure}[h]
    \centering
    \includegraphics[width=0.7\linewidth]{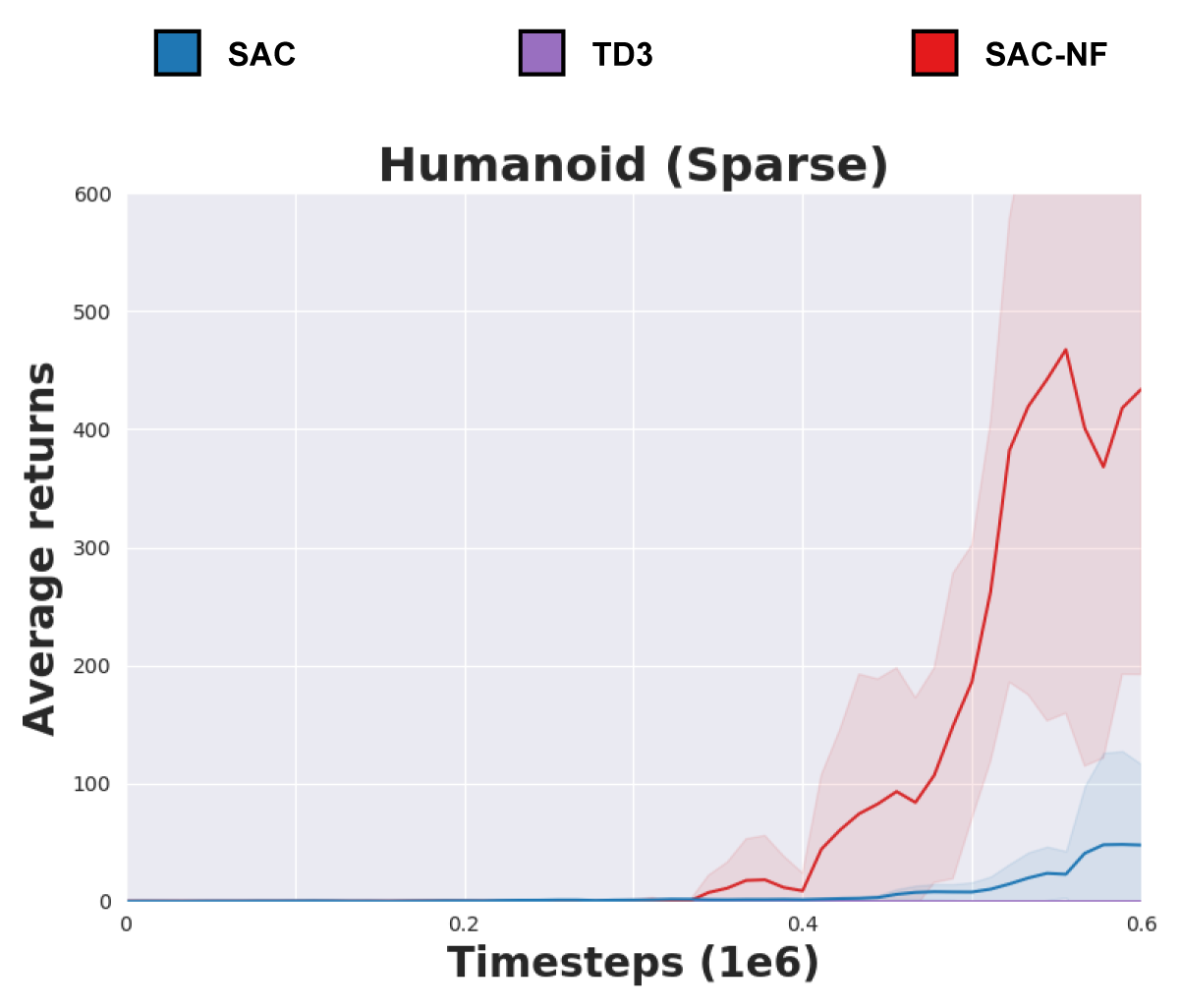}
    \caption{Performance of SAC-NF compared against SAC (Gaussian policy) for a sparse environment in which reward is observed after the agent reaches a certain threshold distance.}
    \label{fig:sparse_envs}
\end{figure}

Even if SAC-NF is meant to better track suboptimal solutions, we tested whether adding normalizing flow layers improves performance within sparse reward environments. To do so, we evaluated on Sparse Humanoid (SparseHU). For SparseHU, a reward of +1 is granted when the agent reaches a distance threshold above $0.6$. As shown in Figure~\ref{fig:sparse_envs}, SAC-NF has better performance than its SAC counterpart and TD3 which struggle to take off.

